\documentclass[sigconf]{acmart}

\renewcommand\footnotetextcopyrightpermission[1]{}
\usepackage{multicol}
\usepackage{multirow}
\usepackage{graphicx}
\usepackage{subfigure}
\usepackage{makecell}
\usepackage{amsmath,mathtools}
\usepackage{color}
\usepackage{algorithm}
\usepackage{algorithmic}
\usepackage{bbold}
\usepackage{colortbl}

\usepackage{bbding}

\AtBeginDocument{%
  \providecommand\BibTeX{{%
    \normalfont B\kern-0.5em{\scshape i\kern-0.25em b}\kern-0.8em\TeX}}}

\begin{document}

\title{Prescribing the Right Remedy: Mitigating Hallucinations in Large Vision-Language Models via Targeted Instruction Tuning}

\author{Rui Hu\textsuperscript{1},\ Yahan Tu\textsuperscript{1},\ Shuyu Wei\textsuperscript{1},\ Dongyuan Lu\textsuperscript{2},\ Jitao Sang\textsuperscript{1}}
\affiliation{%
  \institution{\textsuperscript{1}Beijing Key Lab of Traffic Data Analysis and Mining, Beijing Jiaotong University}
  \country{}
}
\affiliation{%
  \institution{\textsuperscript{2}School of Information Technology and Management, University of International Business and Economics}
  \country{}
}
\email{rui.hu@bjtu.edu.cn}

\begin{abstract}
\urlstyle{tt}

Despite achieving outstanding performance on various cross-modal tasks, current large vision-language models (LVLMs) still suffer from hallucination issues, manifesting as inconsistencies between their generated responses and the corresponding images. Prior research has implicated that the low quality of instruction data, particularly the skewed balance between positive and negative samples, is a significant contributor to model hallucinations. Recently, researchers have proposed high-quality instruction datasets, such as LRV-Instruction, to mitigate model hallucination. Nonetheless, our investigation reveals that hallucinatory concepts from different LVLMs exhibit specificity, i.e. the distribution of hallucinatory concepts varies significantly across models. Existing datasets did not consider the hallucination specificity of different models in the design processes, thereby diminishing their efficacy in mitigating model hallucination. In this paper, we propose a targeted instruction data generation framework named \textit{DFTG} that tailored to the hallucination specificity of different models. Concretely, \textit{DFTG} consists of two stages: hallucination diagnosis, which extracts the necessary information from the model's responses and images for hallucination diagnosis; and targeted data generation, which generates targeted instruction data based on diagnostic results. The experimental results on hallucination benchmarks demonstrate that the targeted instruction data generated by our method are more effective in mitigating hallucinations compared to previous datasets.
\end{abstract}





\maketitle

\section{Introduction}

\begin{figure}
    \centering
    \includegraphics[width=0.85\linewidth]{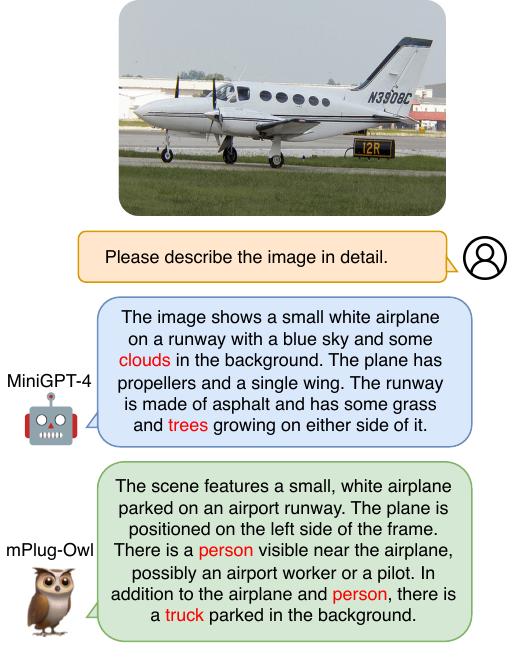}
    \caption{Example of hallucinations in LVLMs. Given an image, the LVLMs output corresponding responses which include objects that do not exist in the image (hallucination object highlighted with red color).}
    \label{fig:幻觉示例}
\end{figure}

The rapid development of large language models (LLMs)~\cite{touvron2023llama, brown2020language} in recent years marks a significant step towards achieving artificial general intelligence (AGI). Recent studies have been devoted to introducing the powerful capabilities of LLMs to the field of multimodal models. Empowered by LLMs, large vision-language models (LVLMs)~\cite{zhu2023minigpt, ye2023mplug, liu2023llava, alayrac2022, li2023a} have demonstrated strong performance on multiple cross-modal tasks, such as visual question answering, and image captioning. Despite the exciting breakthrough, LVLMs suffer from hallucination issues inevitably~\cite{liu2024survey}, which impede their real-world utility. In the field of vision-language models, the term ‘hallucination’ refers to the phenomenon that models generate responses that are inconsistent with the given images. Taking Fig.~\ref{fig:幻觉示例} as an example, the generated content of the LVLMs includes objects (e.g., clouds, person) that are not present in the image.

Several efforts have been made to mitigate hallucinations in LVLMs. Generally, these methods can be divided into two categories: revision-based methods~\cite{yin2023woodpecker, zhou2023analyzing, wu2024logical} and finetuning-based methods~\cite{liu2023mitigating, hu2023ciem, yu2023hallucidoctor, wang2023vigc}. 
Revision-based methods use external experts or self-reflection to correct the generated text during the inference stage. For example, Woodpecker~\cite{yin2023woodpecker} first extracts information from the images and then utilizes ChatGPT~\cite{achiam2023gpt} rewriting generated content to correct hallucinations. Although these methods do not require additional training, they incur heavy inference burdens due to the need for extra inference steps and API calls.

Finetuning-based methods aim to address hallucination issues by tuning models with high-quality instruction data.
On the one hand, as revealed by ~\cite{li2023evaluating}, current LVLMs tend to answer "Yes" for any questions presented to the model, regardless of their accuracy. This tendency stems from their fine-tuning on datasets that predominantly feature positive instructions, lacking a balanced representation of negative ones~\cite{liu2023mitigating}. To address the unbalance problem, Liu et al.~\cite{liu2023mitigating} propose \textit{LRV-Instruction}, a dataset includes both negative and positive instructions for robust instruction tuning. 
On the other hand, ~\cite{yu2023hallucidoctor} reveals that the instruction finetuning data used in the current LVLMs itself contains errors (i.e. texts incongruent with the images), which compromise the LVLMs’ ability to perceive the real world accurately. They then proposed a framework to correct errors in existing instruction tuning datasets~\cite{yu2023hallucidoctor}.

In this paper, we find experimentally that hallucinations from different LVLMs exhibit specificity, making it difficult for the existing instruction data to cover all potential hallucinations of the model. The hallucination specificity refers to the phenomenon, where models exhibit varied hallucination patterns for identical images, due to their unique associations with the depicted scenes (see Sec~\ref{sec:3.2}). 
Taking Fig.~\ref{fig:幻觉示例} as an example, for the same image depicting an airplane parked on a runway, MiniGPT-4~\cite{zhu2023minigpt} associates the \textit{sky} with the non-exist \textit{cloud}, while mPlug-Owl~\cite{ye2023mplug} associates the \textit{airplane} with  the non-exist \textit{person} and \textit{truck}. Considering that different LVLMs utilize different data during the training stage, this hallucination specificity should be quite common in these models.

\begin{figure}[!t]
\setlength{\abovecaptionskip}{0.5em} 
    \centering
    \subfigure[LRV-Instruction]{
    \includegraphics[width=1\linewidth]{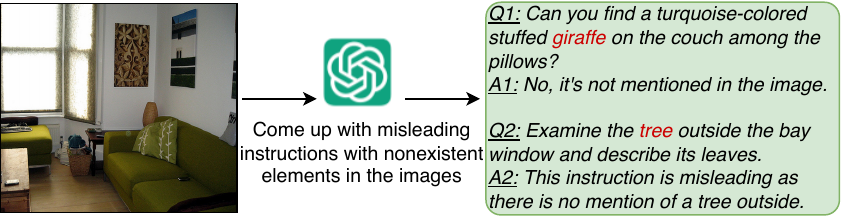}
    }
    \subfigure[DFTG]{
    \includegraphics[width=1\linewidth]{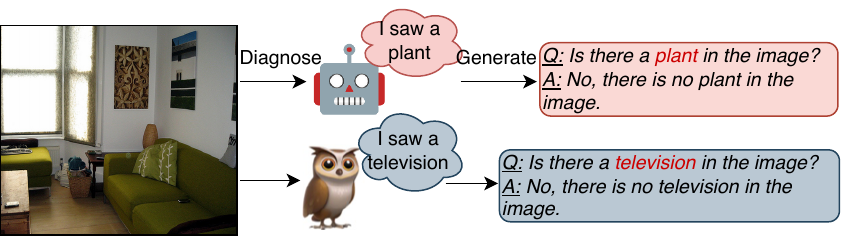}
    }
    \caption{Illustration of the data construction process of LRV-Instruction and our proposed DFTG. Given an image, (a) LRV-Instruction directly uses GPT4 to generate instruction data, while (b) DFTG first diagnoses hallucinations in different LVLMs, then generates targeted instruction data.}
    \label{fig:方法对比示意图}
\end{figure}

However, the existing instruction tuning datasets for mitigating hallucinations do not take into account the hallucination specificity of the LVLMs. For example, as shown in Fig.~\ref{fig:方法对比示意图}(a), LRV-Instruction~\cite{liu2023mitigating} directly uses the in-context few-shot learning ability of GPT4~\cite{achiam2023gpt} to generate instruction data. The problem with this approach lies in the fact that the generated instruction data represent the cognition of GPT4, which may not be applicable to other LVLMs. Based on this, we raize the following question: \textbf{Would considering the hallucination specificity of the LVLMs in the generation of instruction data lead to a more effective mitigation of model hallucinations?}

To answer the qusetion, we propose DFTG (\textbf{D}iagnose \textbf{F}irst, \textbf{T}hen \textbf{G}enerate) framework to generate instruction data based on hallucinations of LVLMs. As illustrated in Fig.~\ref{fig:方法对比示意图}(b), for a given image, DFTG first diagnoses what hallucinations the model perceives in the image and then generates instruction data based on the diagnostic results. Concretely, our framework consists of two stages, the first stage diagnose the model's hallucinations on a given image, including four steps: (1) \textit{Caption generation} obtains the model's description of the image, which includes the perception and association of the model for this image; (2) \textit{text information extraction} identifies the key objects mentioned in the generated sentences; (3) \textit{image information extraction} detects the visual objects in the given image; (4) \textit{ hallucination check} compares the extracted textual and image information to diagnosis hallucinatory objects. The second stage generates positive and negative instructions based on diagnostic results from stage one. The generated dataset is then used to finetune the LVLMs.

We evaluate the effectiveness of our method through extensive experiments on POPE~\cite{li2023evaluating}, MME~\cite{fu2023mme}, AMBER~\cite{wang2023llm} and VHTest~\cite{huang2024visual} datasets. The results indicate the effectiveness of this new paradigm, that is, using targeted instruction data for hallucination mitigation.

In summary, our contributions are as follows:
\begin{itemize}
    \item We find that LVLMs exhibit hallucination specificity, which attenuates the effectiveness of existing instruction tuning datasets in mitigating hallucinations.
    \item We propose the DFTG framework for automatically generating targeted instruction data, which first diagnoses model hallucinations and then generates data.
    \item We perform extensive experiments with hallucination benchmarks, demonstrating the effectiveness of targeted instruction tuning.
\end{itemize}

\begin{table*}
    \caption{The model architectures and datasets employed during the training phases of MiniGPT-4 and mPlug-Owl.}
    \begin{tabular}{ccccc}
    \toprule
        Model     & Vision encoder & Language model & Pre-training dataset & Instruction tuning dataset \\ \midrule
        MiniGPT-4~\cite{zhu2023minigpt} & ViT-L/14            & Vicuna         & \makecell{Conceptual Caption,\\SBU, LAION}           & MiniGPT-3.5k                      \\ \midrule
        mPlug-Owl~\cite{ye2023mplug} & ViT-G/14            & LLaMA          & \makecell{Conceptual Caption,\\COYO, LAION, MSCOCO} & \makecell{Alpaca, Vicuna, Baize, \\LLaVA-150k} \\
    \bottomrule
    \end{tabular}
    \label{tab:model-intro}
\end{table*}

\section{Related Works}

\subsection{Large Vision-Language Models}

With the success of Large Language Models (LLMs), there has been a surge in efforts to incorporate additional modalities into LLMs, leading to the development of Large Vision-Language Models, namely LVLMs~\cite{alayrac2022, li2023a, liu2023llava, chen2023, huang2024, ye2023mplug, zhu2023minigpt}. Typically, LVLMs consist of a visual encoder and a language model connected via a connector. Through training on image-text pairs, LVLMs achieve robust visual understanding and accomplish multimodal tasks such as Visual Question Answering (VQA) and image captioning.

Early models mainly focus on large-scale image-text pre-training. Subsequently, Visual instruction tuning~\cite{liu2023llava} rapidly emerged as a prominent training paradigm in the multimodal domain. For example, InstructBLIP~\cite{dai2023} enhances the pre-trained BLIP-2~\cite{li2023a} by training on instruction tuning datasets. LLaVA~\cite{liu2023llava} further strengthens the model's instruction-following capabilities by training on visual instruction-following data. 


\subsection{Halluciations In LVLMs}

Despite the excellent cross-modal understanding capabilities of LVLMs, they suffer from hallucination issues similar to those of LLMs. Existing methods for mitigating  hallucinations of LVLMs can be categorized into revision-based and finetuning-based methods, depending on the the stage of hallucination processing (i.e. inference phase and training phase).

The revision-based methods aims to remove hallucinations after the model generates a response. To acheve this, they usually need to make use of additional expert models. For example, 
Woodpecker~\cite{yin2023woodpecker} with the help of additional models such as an VQA model and ChatGPT to locate hallucinatory sentence and rewrite them to remove the hallucinations.
LURE~\cite{zhou2023analyzing} trained an additional model for receiving hallucinatory input and reconstructing it to eliminate hallucinations. In summary, this type of methods first require the original model to generate a response and then fix it on top of that, which increases the time and monetary cost of inference.

The finetuning-based methods use relatively less data and to tune the model for better alignment of images and texts. Compared to the revision-based methods, this type of methods adds no additional inference costs. For instance, LRV-Instruction~\cite{liu2023mitigating} introduces a comprehensive instruction-following data that includes both positive and negative instructions to address the problem of unbalanced data distribution.
Hu et al.~\cite{hu2023ciem} propose CIEM framwork to automatically
construct contrastive question-answer pairs for evaluating and
funetuning the models. Yu et al.~\cite{yu2023hallucidoctor} proposed the HalluciDoctor framework to correct errors in instruction tuning datasets which compromise the LVLMs’ ability to perceive the real world accurately. 

In contrast to these methods, our method starts from the model itself, constructing targeted instruction tuning data to mitigate hallucinations of different LVLMs.

\section{Preliminaries}

This section describes the details of the analytical experiments in which we observe the hallucination specificity of different LVLMs. In Section 2.1, we first introduce how instruction tuning is used to mitigate hallucinations in LVLMs. We then show the hallucination specificity of LVLMs in Section 2.2. In Section 2.3, we demonstrate the inadequacy of using GPT4-generated instruction data to mitigate hallucinations.

\subsection{The Evolution of Instruction Tuning in LVLMs}

The training process of current LVLMs typically consists of two stages: pre-training stage and instruction tuning stage. During the pre-training stage, the model is designed to acquire vision-language knowledge from a large collection of aligned image-text pairs~\cite{zhu2023minigpt}. After the pre-training stage, although the model has acquired a significant amount of knowledge, it is unable to generate meaningful output for user requests. The purpose of instruction tuning is to enable the model to generate reliable responses under different user instructions. Compared to the pre-training stage, instruction tuning typically utilizes small-scale, high-quality datasets. 

A common practice for generating instruction data is to take advantage of the ability of a more powerful model to generate it automatically. For example, Liu et al.~\cite{liu2023llava} leverage language-only GPT4 as the strong teacher to convert image-caption pairs into instruction-following data (i.e. LLaVA-150k) involving visual content. Although instruction data has higher quality compared to pre-training data, there are still certain issues present, such as an imbalanced distribution of positive and negative samples. These issues have been proven to be one of the causes leading to model hallucinations~\cite{li2023evaluating, liu2023mitigating}. Therefore, researchers continue to improve the quality of instruction data, hoping to mitigate model hallucination during the instruction tuning stage~\cite{liu2023mitigating, hu2023ciem}. Liu et al.~\cite{liu2023mitigating} have taken a significant step in this direction, they create LRV-Instruction which is the first large visual instruction tuning dataset for mitigating model hallucinations. The construction process of LRV-Instruction is similar to LLaVA-150k, which also utilizes the capabilities of GPT4 to generate instruction data. 

One key reason why LRV-Instruction can be used to alleviate model hallucination is that it has negative samples in addition to positive ones. As shown in Fig.~\ref{fig:方法对比示意图}(a), the instructions for negative samples are misleading, as they contain objects that do not exist in the image. Models finetuned on such data can learn to respond "No" to misleading instructions.

\subsection{LVLMs Exhibit Hallucination Specificity}
\label{sec:3.2}

To better understand how instruction data alleviates model hallucinations, we first analyzed the hallucinations of different LVLMs. We follow the choice of Liu et al.~\cite{liu2023mitigating} which uses the MiniGPT-4~\cite{zhu2023minigpt} and mPlug-Owl~\cite{ye2023mplug} as the backbone models. Table ~\ref{tab:model-intro} shows the similarities and differences between the two models in terms of model architecture and dataset selection. They both use ViT~\cite{dosovitskiy2020image} (vison transfromer) as a vision encoder and LLaMA-based language model (Vicuna~\cite{vicuna2023} is a variant of LLaMA~\cite{touvron2023llama}). However, they differ in the selection of training data, especially in the instruction tuning data. 

\begin{figure*}[t]
\setlength{\abovecaptionskip}{0em} 
    \centering
    \subfigure[The top 20 most frequent hallucinatory objects on all images]{
        \includegraphics[width=0.24\textwidth]{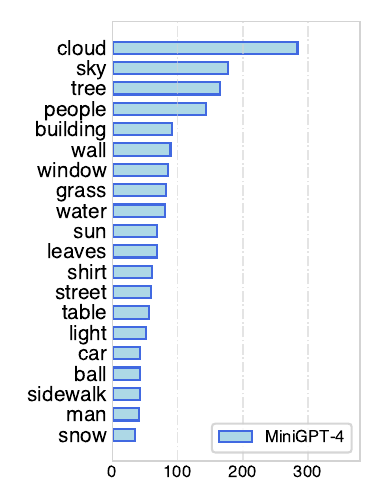}
        \includegraphics[width=0.24\textwidth]{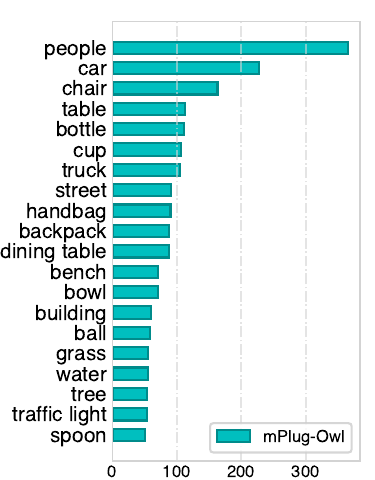}
    }
    \subfigure[The top 20 most frequent hallucinatory objects on "kitchen" scence images]{
        \includegraphics[width=0.24\textwidth]{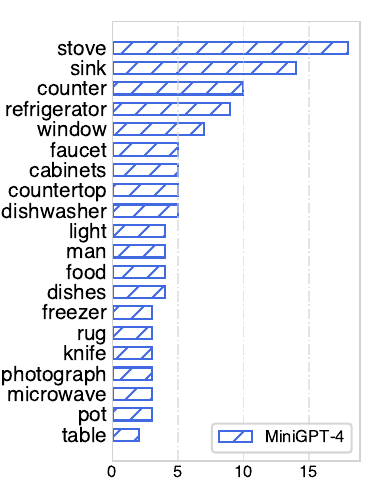}
        \includegraphics[width=0.24\textwidth]{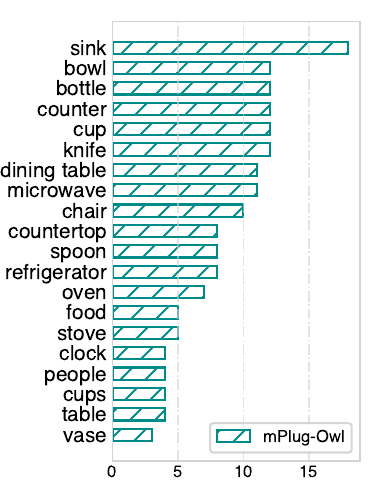}
    }
    \caption{Frequency statistics of hallucinatory objects in MiniGPT-4 and mPlug-Owl.}
    \label{fig:幻觉频率}
\end{figure*}

\begin{table}[!tbh]
    \caption{Similarity of the hallucinatory object sets of MiniGPT-4 and mPlug. TopK represents taking the top K elements from the object sets.}
    \centering
    \begin{tabular}{ccccc}
    \toprule
        \multirow{2}{*}{TopK} & \multicolumn{2}{c}{All images} & \multicolumn{2}{c}{"kitchen" images} \\ \cline{2-5}
        
        & Overlap & RBO value & Overlap & RBO value \\ \midrule
        
        @5 & 20.0\% & 0.090 & 40.0\% & 0.347 \\
        @10 & 10.0\% & 0.109 & 30.0\% & 0.312 \\
        @15 & 26.7\% & 0.132 & 40.0\% & 0.319 \\
        @20 & 45.0\% & 0.212 & 45.0\% & 0.348 \\
        
    \bottomrule
    \end{tabular}
    \label{tab:overlap}
    \vspace{0em}
    
\end{table}

We randomly select 2,000 images from the COCO2017 dataset~\cite{lin2014microsoft} that contain diverse scenes and objects. We then detect and count the hallucinatory objects of MiniGPT-4 and mPlug-Owl in these images (the method of detecting hallucinatory objects will be described in detail in Section~\ref{sec:4}). The statistical results are shown in Fig.~\ref{fig:幻觉频率}. Furthermore, to compare the differences between the model hallucinations more clearly, we use the overlap ratio and RBO\footnote{RBO compares two ranked lists, and returns a numeric value between zero and one to quantify their similarity. A RBO value of zero indicates that the lists are completely different, and a RBO value of one means completely identical.} (Rank-biased overlap) to measure the similarity of the two sets of hallucinatory objects. The calculation results are shown in Table~\ref{tab:overlap}.

~\\
\noindent\textbf{The frequency distribution of hallucinatory objects varies across LVLMs.} We conduct a frequency statistics on the hallucinatory objects generated by the two models and find significant differences between them. As shown in Fig.~\ref{fig:幻觉频率}(a), the top 20 most frequent hallucinatory objects in MiniGPT-4 and mPlug-Owl are not identical, for example, the most frequent hallucinations of MiniGPT-4 are "cloud" and "sky", while the ones of mPlug-Owl are "people" and "car".
As shown in Table~\ref{tab:overlap}, there is some overlap in the hallucinatory object sets of the two models (such as "people", "building"). This is reasonable because these objects are very common and appear frequently inside the training data of both models, causing the model hallucination. However, the RBO value of the two hallucinatory object sets remains relatively low, indicating the difference in the frequency distribution of hallucinatory objects between the two models.

\begin{table}[!tbh]
    \caption{Similarity of the hallucinatory object sets of LVLMs and non-exist object set generated by GPT4.}
    \centering
    \begin{tabular}{ccccc}
    \toprule
        \multirow{2}{*}{TopK} & \multicolumn{2}{c}{\makecell{MiniGPT-4 \& GPT4}} & \multicolumn{2}{c}{\makecell{mPlug-Owl \& GPT4}} \\ \cline{2-5}
        
        & Overlap & RBO value & Overlap & RBO value \\ \midrule
        
        @5 & 20.0\% & 0.090 & 0.0\% & 0.0 \\
        @10 & 40.0\% & 0.200 & 10.0\% & 0.021 \\
        @15 & 46.7\% & 0.275 & 20.0\% & 0.070 \\
        @20 & 40.0\% & 0.307 & 20.0\% & 0.106 \\
        
    \bottomrule
    \end{tabular}
    \label{tab:overlap2}
\end{table}

~\\
\noindent\textbf{Different LVLMs have different associations for the same scene.} One of the reasons for hallucinations is that models learn spurious correlations from training data~\cite{li2023evaluating}. We find that different LVLMs learn different spurious correlations. As shown in Fig.~\ref{fig:幻觉频率}(b), we count the hallucinations of MiniGPT-4 and mPlug-Owl on the images of the "kitchen" scene. In addition to differences in frequency distributions, we find that their hallucinations represent their underlying perceptions of the scene. 
To confirm this, we let ChatGPT summarize the similarities and differences between the two lists of hallucinated objects (named list A and list B). The answer from ChatGPT is as follows: "\textit{Both lists primarily focus on kitchen items, but list B seems to have a broader scope, including items related to dining and dining spaces, while list A appears to be more focused on kitchen infrastructure and appliances.}", indicating that different models do have different associations for the same scene.

\begin{figure*}
\setlength{\abovecaptionskip}{0.5em} 
    \centering
    \includegraphics[width=1\textwidth]{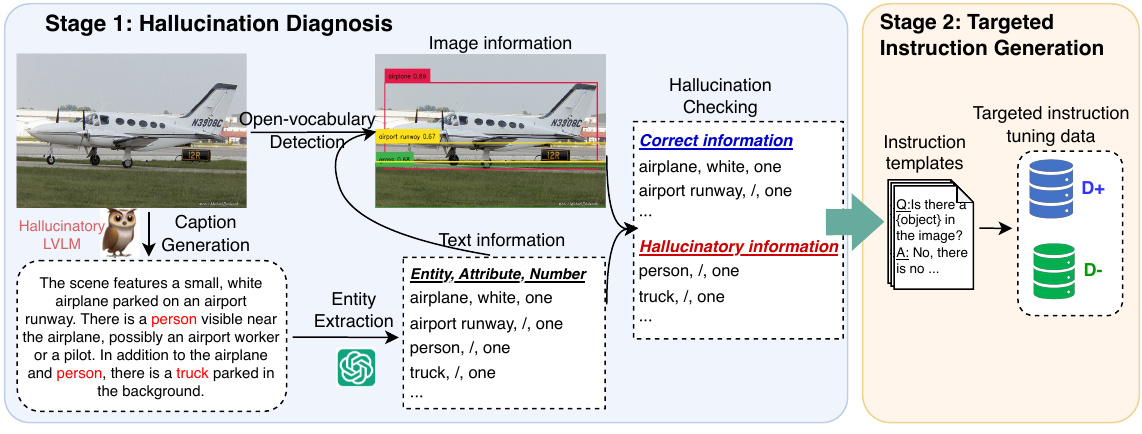}
    \caption{Illustration of our framework. The first stage extracts the information needed for hallucination diagnosis from images and textual description, the second stage diagnoses hallucinations and generates targeted data.}
    \label{fig:pipeline}
\end{figure*}

\begin{figure*}
\setlength{\abovecaptionskip}{0.5em} 
    \centering
    \includegraphics[width=0.98\textwidth]{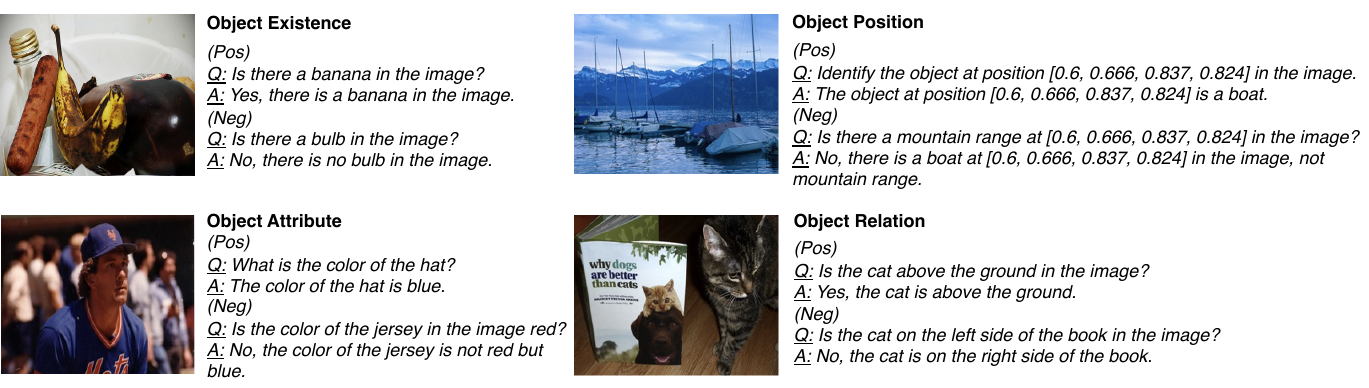}
    \caption{Instances of four different types of instruction data generated by our framework DFTG. \textit{(Pos)} and \textit{(Neg)} represent positive and negative samples, respectively.}
    \label{fig:instances}
    
\end{figure*}

\subsection{GPT4-generated Instruction Data Lacks Specificity}

Since the instruction tuning datasets represented by LRV-Instruction use GPT4 to generate positive and negative samples, we then analyze the differences between GPT4-generated non-existing objects~\footnote{The top 20 most frequent non-existent objects in LRV-Instruction are \textit{[man, dog, woman, tree, cat, umbrella, sky, flower, car, tennis, train, bicycle, table, building, hat, bird, ball, elephant, ballcon, window, clock, baseball, boy]}} in negative samples and hallucinatory objects of LVLMs. For a fair comparison, we select a subset of images from LRV-Instruction and count hallucinatory objects of MiniGPT-4 and mPlug-Owl on these images. As shown in Table~\ref{tab:overlap2}, the overlap and RBO value of the hallucinatory object sets of LVLMs and non-exist object set generated by GPT4 are relatively low. One reason for this is that GPT4 is divergent when generating negative samples. Take Fig.~\ref{fig:方法对比示意图}(a) as an example, for an indoor scene, GPT4 generates a negative sample about \textit{"giraffe"}, which is less relevant to the scene, and the LVLMs is less likely to produce this hallucination.

In conclusion, because of the hallucinatory specificity of the LVLMs and the properties of GPT4 itself make GPT4-generated instruction data lacks
specificity.

\section{Method}
\label{sec:4}

In this section, we introduce our DFTG framework consisting of two stages: 1) Hallucination diagnosis (Sec.~\ref{sec:4.1}), which includes four steps: caption generation, text information extraction, image information extraction and hallucination checking. 2) Targeted instruction data generation (Sec.~\ref{sec:4.2}) that generates targeted instruction data based on the diagnosis results. An overview of our framework is depicted in Fig.~\ref{fig:pipeline}.

\begin{table*}
    \centering
    \caption{Results on POPE. The best and second-to-best performances within each setting are \textbf{bolded} and underlined, respectively.}
    \begin{tabular}{cl|cccccc|cc}
    \toprule
       \multirow{2}{*}{Model} & \multirow{2}{*}{Method} & \multicolumn{2}{c}{Adversarial} & \multicolumn{2}{c}{Popular} & \multicolumn{2}{c}{Random} & \multicolumn{2}{c}{\makecell{Average}}  \\ \cline{3-10}
       & & Accuracy & F1-Score & Accuracy & F1-Score & Accuracy & F1-Score & Accuracy & F1-Score \\ \midrule
       \multirow{4}{*}{MiniGPT-4} 
       & Vanilla            & 51.40 & 62.83 & 51.73 & 64.90 & 55.20 & 68.75 & 52.78 & 65.49 \\
       & +LRV-Instruction    & \underline{66.87} & \underline{64.07} & 68.73 & \underline{67.85} & 73.43 & 74.23 & 69.68 & \underline{68.72} \\
       & +VIGC               & 66.20 & 55.57 & \underline{70.70} & 63.72 & \underline{77.63} & \underline{74.56} & \underline{71.51} & 64.62   \\
       \rowcolor{gray!10}& +Ours               & \textbf{73.53} & \textbf{76.00} & \textbf{75.87} & \textbf{78.59} & \textbf{78.03} & \textbf{80.98} & \textbf{75.81} & \textbf{78.52}  \\ \midrule
       
       \multirow{4}{*}{mPlug-Owl} 
       & Vanilla            & 54.63 & 29.35 & 58.03 & 37.56 & 62.63 & 48.37 & 58.43 & 38.43 \\
       & +LRV-Instruction    & \underline{61.20} & \underline{53.46} & \underline{66.10} & \underline{61.80} & \underline{68.50} & \underline{68.60} & \underline{65.27} & \underline{61.28} \\
       & +VIGC               & 54.57 & 23.02 & 57.13 & 31.29 & 61.43 & 41.88 & 57.71 & 32.06 \\
       \rowcolor{gray!10} & +Ours               & \textbf{63.03} & \textbf{66.24} & \textbf{66.57} & \textbf{69.85} & \textbf{71.33} & \textbf{75.09} & \textbf{66.98} & \textbf{70.39} \\ 
    \bottomrule
    \end{tabular}
    \label{tab:pope}
\end{table*}

\subsection{Hallucination Diagnosis}
\label{sec:4.1}

Analyzing the potential hallucinations that may exist in an LVLM on an image requires two types of information: on the one hand, the perceptual information of the model towards the image; on the other hand, the true information of the image. In this section, we first extract these two aspects of information and then diagnose hallucinations using these information.

~\\
\textbf{Caption generation.} The description of a given image generated by a model demonstrates how it perceives the image. Therefore, the first step is to collect descriptions of the images. Specifically, we use "\textit{Describe the image in detail.}" as the prompt to let the LVLMs generate detailed descriptions.

~\\
\textbf{Text information extraction.} After obtaining an image description, we need to extract the key information (e.g., objects, their attributes and quantities) for subsequent processing. For example, given a sentence "\textit{The scene features a white airplane}", we want to extract the included triplet \textit{\{object, attribute, quantity\}}: \textit{\{airplane, white, one\}}. Since LLMs have been proven to possess the in-context learning ability~\cite{brown2020language}, which means that by providing a few task examples, the model can learn how to perform the task, we prompt an LLM to complete this task.


~\\
\noindent\textbf{Image information extraction.} After obtaining the key information in the description of an image (which represents how the model perceives the image), we need to extract the real image information for comparison. Object detection models are often used to determine whether certain objects exist in an image, but they usually require pre-defined object categories. Thanks to the advancement of pre-trained vision-language models (VLMs), such as CLIP~\cite{radford2021learning}, researchers have integrated object detection models with VLMs, endowing them with the capability of open-vocabulary object detection (OVOD)~\cite{kuo2022f, liu2023grounding}. We input the entities extracted from the description into the OVOD model to obtain information about the entities in the image, such as quantity, bounding-box, etc.

~\\
\textbf{Hallucination checking.} After collecting the perceptual information of the model and the ground truth information of the image, the next step involves comparing these pieces of information to identify the hallucinations in the model. To ascertain whether an entity constitutes a hallucination, we conduct a comparison between the entity's quantity mentioned in the description and its quantity as detected in the actual image. If the quantity of the entity metioned in the description exceed zero while no instances of the entity are detected in the image, we classify said entity as hallucinatory.

For object attributes, we consider the attribute-object pair as a unified entity and let the OVOD model detect them. If the quantity of a specific object is non-zero while the quantity of the attribute-object pair is zero in the detection results, we categorize that attribute as hallucinatory.
For example, if the OVOD model detects one "\textit{airplane}" in the given image but does not detect any "\textit{red airplane}", then "\textit{red}" is considered a hallucinatory attribute.

\subsection{Targeted Instruction Data Generation}
\label{sec:4.2}

\begin{table}
    \centering
    \caption{Results on MME Existence subset. The best result is highlighted in boldface.}
    \begin{tabular}{cl|cc|c}
    \toprule
       Model & Method & Acc & Acc+ & Total \\  \midrule
       \multirow{4}{*}{MiniGPT-4} 
       & Vanilla            & 66.67 & 43.33 & 110.00 \\
       & +LRV-Instruction   & 83.33 & 70.00 & 153.33 \\
       & +VIGC              & 73.33 & 46.67 & 120.00 \\
       \rowcolor{gray!10} & +Ours              & \textbf{91.67} & \textbf{83.33} & \textbf{175.00} \\ \midrule
       
       \multirow{4}{*}{mPlug-Owl} 
       & Vanilla            & 61.67 & 26.67 & 88.33 \\
       & +LRV-Instruction   & 71.67 & \textbf{46.67} & 118.33 \\
       & +VIGC              & 66.67 & 36.67 & 103.33 \\
       \rowcolor{gray!10} & +Ours              & \textbf{73.33} & \textbf{46.67} & \textbf{120.00} \\ 
    \bottomrule
    \end{tabular}
    \label{tab:mme}
    \vspace{-1em}
\end{table}

~\\
This stage is responsible for generating instruction tuning data tailored to the model to correct its perception. Specifically, we generate four types of instruction data based on the information extracted from the stage one: object existence, object attribute, object position and object relation.

We have designed instruction templates separately for each of these four types for automatic generation. For example, for object existence, we use "\textit{Is there a \{object\} in the image?}" as an instruction template. Based on whether the \textit{object} is correct or a hallucination, the answer templates are "\textit{Yes, there is a \{object\} in the image.}" and "\textit{No, there is no \{object\} in the image}.", respectively. For object attribute, we use "\textit{Is the \{object\} \{attribute\} in the image?}" as the instruction template and based on where the \textit{\{attribute\}} is correct or hallucinatory to construct the answer.

\begin{table*}
    \centering
    \caption{Results on AMBER. The best and second-to-best performances are \textbf{bolded} and underlined, respectively.}
    \begin{tabular}{cl|cccccc|cc}
    \toprule
       \multirow{2}{*}{Model} & \multirow{2}{*}{Method} & \multicolumn{2}{c}{Existence} & \multicolumn{2}{c}{Attribute} & \multicolumn{2}{c}{Relation} & \multicolumn{2}{c}{\makecell{Descriminative Task}}  \\ \cline{3-10}
       & & Accuracy & F1-Score & Accuracy & F1-Score & Accuracy & F1-Score & Accuracy & F1-Score \\ \midrule
       \multirow{4}{*}{MiniGPT-4} 
       & Vanilla            & 57.31 & 72.82 & 46.78 & 48.40 & 41.41 & \underline{43.89} & 49.80 & 60.08 \\
       & +LRV-Instruction   & \underline{62.31} & \underline{76.73} & 55.89 & \underline{56.84} & 57.45 & 22.91 & \underline{58.29} & \underline{64.94} \\
       & +VIGC              & 33.35 & 49.98 & \textbf{61.93} & 47.90 & \textbf{60.40} & 10.84 & 51.85 & 46.85 \\
       \rowcolor{gray!10} & +Ours              & \textbf{89.28} & \textbf{94.28} & \underline{58.78} & \textbf{66.76} & \underline{58.23} & \textbf{55.92} & \textbf{69.28} & \textbf{78.56} \\ \midrule
       
       \multirow{4}{*}{mPlug-Owl} 
       & Vanilla                    & 21.89 & 35.89 & 52.77 & 30.88 & 54.21 & 35.13 & 42.24 & 33.78 \\
       & +LRV-Instruction           & \underline{48.72} & \underline{65.48} & \underline{54.78} & \underline{56.43} & \underline{54.27} & 43.49 & \underline{52.62} & \underline{59.62} \\
       & +VIGC                      & 20.43 & 33.90 & \textbf{55.05} & 32.20 & \textbf{55.23} & \underline{50.30} & 43.08 & 35.20 \\
       \rowcolor{gray!10} & +Ours   & \textbf{76.54} & \textbf{86.66} & 54.42 & \textbf{60.53} & 44.89 & \textbf{53.56} & \textbf{60.97} & \textbf{71.51} \\ 
    \bottomrule
    \end{tabular}
    \label{tab:amber}
\end{table*}

\begin{table*}
    \centering
    \caption{Accuracy on VHTest. The best and second-to-best performances are \textbf{bolded} and underlined, respectively.}
    \begin{tabular}{cl|cccccccc|c}
    \toprule
       Model & Method & Color & Counting & Existence & OCR & Orientation & Position & Shape & Size & Average \\ \midrule
       
       \multirow{4}{*}{MiniGPT-4} 
       & Vanilla & 41.33 & 44.67 & 50.67 & 36.67 & \textbf{47.33} & 44.00 & 46.67 & 40.00 & 43.92 \\
       & +LRV-Instruction & 46.67 & 29.33 & \textbf{60.67} & 40.67 & 33.33 & 43.33 & 46.67 & 40.67 & 42.67 \\
       & +VIGC & \underline{56.00} & \underline{48.00} & 52.67 & \underline{52.00} & \textbf{47.33} & \underline{54.00} & \textbf{52.67} & \underline{52.67} & \underline{51.92} \\
       \rowcolor{gray!10} & +Ours & \textbf{56.67} & \textbf{54.00} & \underline{60.00} & \textbf{54.00} & 40.00 & \textbf{56.67} & \textbf{52.67} & \textbf{54.00} & \textbf{53.50} \\ \midrule
       
       \multirow{4}{*}{mPlug-Owl}
        & Vanilla & 50.67 & \textbf{44.67} & 48.00 & 44.67 & \textbf{50.00} & 42.67 & 43.33 & 50.00 & 46.75 \\
        & +LRV-Instruction & 42.67 & 41.33 & 53.33 & 40.00 & 39.33 & 42.67 & \textbf{60.67} & 44.67 & 45.58 \\
        & +VIGC & \underline{54.67} & \underline{42.67} & \underline{48.00} & \underline{49.33} & 42.67 & \underline{44.00} & 50.00 & \underline{52.67} & \underline{48.00} \\
        \rowcolor{gray!10} & +Ours & \textbf{58.67} & 33.33 & \textbf{57.33} & \textbf{54.67} & 44.67 & \textbf{52.67} & \underline{51.33} & \textbf{56.67} & \textbf{51.17} \\
    \bottomrule
    \end{tabular}
    \label{tab:vhtest}
\end{table*}

For object position, we utilize the bounding boxes of objects to construct instructions aimed at enhancing the model's perception of positions.  Further, we compute the positional relationship between objects based on the bounding box (e.g., on the left side) and generate the corresponding positive and negative samples. The instances of our instruction data are shown in Fig.~\ref{fig:instances}

\section{Experiments}

\subsection{Experimental Settings}
\textbf{Datasets.} \textbf{POPE}~\cite{li2023evaluating} is dedicated to evaluating object hallucinations of LVLMs. It consists of three evaluation settings: random, popular, adversarial. The main difference between them is the strategy for sampling negative objects. \textit{Random Sampling} randomly sample the objects that do not exist in the image; \textit{Popular Sampling} select the top-k most frequent objects in dataset that do not exist in the image; For \textit{Adversarial Sampling}, objects that most frequently co-occur but do not exist in the image are sampled.

\noindent\textbf{MME}~\cite{fu2023mme} serves as a comprehensive benchmark for evaluating common capabilities of LVLMs across a wide spectrum of tasks. We only the existence subset to evaluate the object-level hallucination. We use accuracy and accuracy+ as metrics, where the former is calculated based on each question, while the latter is based on each image, requiring both questions to be answered correctly.

\noindent\textbf{AMBER}~\cite{wang2023llm} is a multi-dimensional benchmark for LVLMs hallucination evaluation. They collect high-quality images and provide comprehensive annotations to facilitate the hallucination evaluation, covering three types of hallucination: existence, attribute and relation.

\noindent\textbf{VHTest}~\cite{huang2024visual} is a recently released dataset for hallucination evaluation. They first finds some initial hallucination modes from existing image datasets and then uses a text-to-image generative model to generate images based on the text descriptions.

~\\
\noindent\textbf{Baselines.} Follow Liu et al.~\cite{liu2023mitigating}, we choose two MiniGPT-4~\cite{zhu2023minigpt} and mPlug-Owl~\cite{ye2023mplug} as our backbone models. We use vanilla models, LRV-Instruction~\cite{liu2023mitigating} and VIGC~\cite{wang2023vigc} as our baselines, where LRV-Instruction and VIGC are instruction datasets that have been proposed to mitigate hallucinations. We generate an 79K-instruction data for MiniGPT-4 and an 58K-instruction data for mPlug-Owl based on the 2,000 COCO2017 images mentioned in section~\ref{sec:3.2}.

~\\
\noindent\textbf{Implementation details.} Follow Woodpecker~\cite{yin2023woodpecker}, we choose GPT-3.5-turbo~\cite{achiam2023gpt} to fulfill the task of text information extract, and use Grounding DINO~\cite{liu2023grounding} as the OVOD model to extract image information. As for MiniGPT-4, we instruct-tune the model with the linear projection layer as the only learnable module. As for mPLUG-Owl, we train the text encoder by LoRA training. We trained the 7B version models on NVIDIA RTX 3090.

\subsection{Main Results}

\textbf{Results on POPE and MME.} Table~\ref{tab:pope} and Table~\ref{tab:mme} show quantitative experimental results on POPE and MME Existence subset, respectively. The models after instruction tuning using the data generated by our framework outperforms the original model and the baseline method. For POPE, our model consistently shows performance improvements under different settings.
Notably, all models have the lowest performance with the Adversarial setting and the highest in the Random setting, indicates that the models learn spurious correlations in training data. The performance gap under different settings of our method is relatively small, demonstrate that our targeted data corrects the model's spurious correlations to some extent. For MME Existence subset, our method results in increases in both accuracy and accuracy+, demonstrating the effectiveness of our method in mitigating object existence hallucination.
\begin{figure}[!thb]
\setlength{\abovecaptionskip}{0.3em} 
    \centering
    \subfigure[POPE Adversarial]{
        \includegraphics[width=0.5\linewidth]{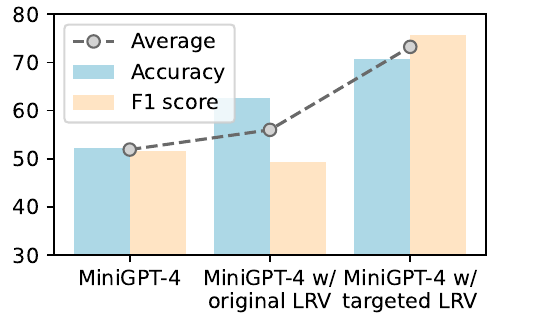}
    }
    \hspace{-1.5em}
    \subfigure[POPE Popular]{
        \includegraphics[width=0.5\linewidth]{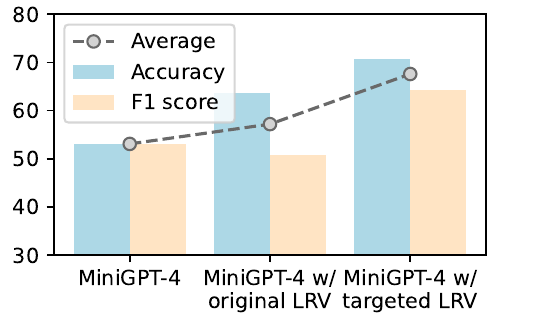}
    }
    \subfigure[POPE Random]{
        \includegraphics[width=0.5\linewidth]{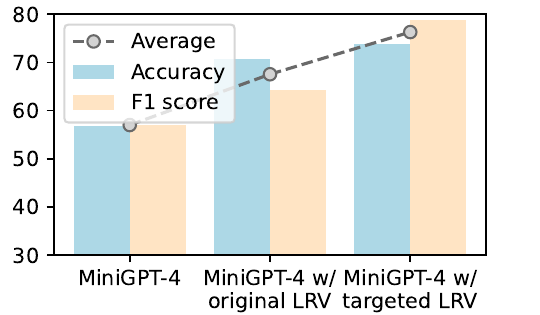}
    }
    \hspace{-1.5em}
    \subfigure[AMBER Discriminative]{
        \includegraphics[width=0.5\linewidth]{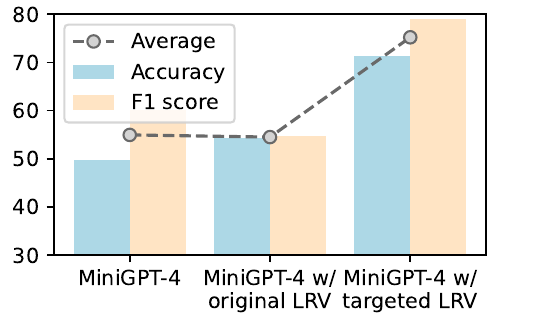}
    }
    \caption{Results comparison on POPE and AMBER between original LRV finetuned MiniGPT-4 and targeted LRV finetuned MiniGPT-4.}
    \label{fig:效果对比}
    \vspace{-1em}
\end{figure}

~\\
\noindent\textbf{Results on AMBER.} Apart from object existence hallucination evaluation in POPE and MME, we used AMBER to evaluate hallucinations from more dimensions. The experimental results of AMBER are shown in Table~\ref{tab:amber}. Firstly, the performance of our method on Existence is significantly outperformed other baselines. Secondly, for Attribute and Relation, our method achieve the best F1 score. 
We find that VIGC achieve the best accuracy but with a low F1 score. After observing responses of VIGC on AMBER questions, we find that it still remain "Yes" bias, for example, the yes ratio of MiniGPT-4 funetuned with VIGC on AMBER is up to 75.1\%, resulting in a low F1 score. Overall, our method has the largest improvement for the original model on the discriminant task (Existence, Attribute, Relation as a whole).

~\\
\noindent\textbf{Results on VHTest.} VHTest uses the text-to-image model to generate eight types of difficult samples, including Color, Counting, etc. The results on VHTest is shown at Table~\ref{tab:vhtest}. Our method achieve comparable performance on the three types of test data, Color, Existence and Position, because our generated data include these types. To our surprise, in some types not covered by our data, such as OCR and Size, our method also shows improvement compared to the original model and other methods. This indicates that our targeted instruction data can improve the common perception capacity of the model besides mitigating hallucinations.

\subsection{Further Analysis} 

\textbf{Robustness on image sources.} To verify the robustness of our method across image sources and further compare it with LRV-Instruction, we conduct comparative experiments under unified settings. We select a subset of images from LRV-Instruction containing 1,000 positive and 1,000 negative samples, denoted as the original LRV dataset. Using these same images, we generate 1,000 positive and 1,000 negative samples tailored for MiniGPT-4, denoted as targeted LRV.

We fine-tune MiniGPT4 with these two datasets separately and evaluate them on POPE and AMBER. As shown in Fig.~\ref{fig:效果对比}, with consistent image source and data size, the targeted LRV tuned MiniGPT-4 outperforms the original MiniGPT-4 and the original LRV tuned MiniGPT-4. This demonstrates the robustness of our method across image sources. It also further demonstrates the effectiveness of generating targeted instruction data for hallucination mitigation.

~\\
\noindent\textbf{Effects of different types of instruction data.} We further compare the effect of using only the data of the object existence type to using all four  types of data. As shown in Table~\ref{tab:only}, using only object existence type data yields better results than the original model. However, incorporating all four types instruction further enhances model performance, indicating the necessity of increasing the diversity of instruction data.

\begin{table}[]
    \centering
    \caption{F1 score comparison between only use object
existence data and use all four types instruction data.}
    \begin{tabular}{cc|ccc}
    \toprule
    \multirow{1}{*}{Dataset} & \multirow{1}{*}{Setting} & \multicolumn{1}{c}{\makecell{Original\\MiniGPT-4}} & {\makecell{Only Object\\Existence}} & \multicolumn{1}{c}{\makecell{All Four\\Types}} \\ \midrule

     \multirow{3}{*}{POPE} 
        & Adversarial & 62.83 & 71.65 & \textbf{75.77} \\
        & Popular & 64.90 & 72.36 & \textbf{76.29} \\
        & Random & 68.75 & 75.99 & \textbf{78.84} \\ \midrule
    \multirow{3}{*}{AMBER} 
        & Existence & 72.82 & 87.29 & \textbf{93.16} \\
        & Attribute & 48.40 & 63.08 & \textbf{64.22} \\
        & Relation & 43.89 & 52.40 & \textbf{53.28} \\
    
    \bottomrule
    \end{tabular}
    \label{tab:only}
    \vspace{-1em}
\end{table}

\section{Discussion}

\textbf{Limitations.} We provide a preliminary scheme for generating targeted instruction tuning data to mitigate hallucinations of LVLMs, wherein there are some limitations. Firstly, our method is affected by the capabilities of the models used as information extractors, such as the ability of ChatGPT to find information about entities in text and the detection capabilities of the OVOD model. The data quality is tied to the accuracy of the information they provide. 

Another limitation is the types of hallucinations that can be diagnosed. Currently, we can only diagnose object and attribute-level hallucinations, but not more complex ones such as action and relation-level hallucinations. This limits the diversity of instruction data and the effect of repairs.

~\\
\noindent\textbf{Future works.} In addition to address the limitations, we believe that the following directions can be explored in the future. The first is the training methods of instruction tuning. Current works optimize the model individually for each sample; however, the model's accurate perception and hallucinations correlated. We can introduce additional training mechanisms, such as contrastive learning~\cite{chen2020simple}, to leverage this correlation. 

Iterative repair is another direction worth exploring. Although current methods can partially mitigate hallucinations, they cannot guarantee complete repair or ensure the absence of new hallucinations. We believe that iteratively diagnosing and repairing is one approach to addressing this issue.

\section{Conclusion}

We propose the DFTG framework, which first diagnoses the hallucinations of LVLMs and then generates targeted instruction data for the model tuning. Compared to previous methods utilizing GPT4 to generate instruction data, our method takes into account the hallucination specificity of different models. Experimental results on hallucination benchmarks indicate the effectiveness of our method in hallucination mitigation.

\bibliographystyle{ACM-Reference-Format}
\balance
\bibliography{sample-base}

\end{document}